# Selection Policies for Equal-Byte Mixed-Precision Quantization of Small Language Models

Equal-Byte Mixed-Precision Search with Random Sampling, Tournament Selection, and Alternis

**Smit Anand | Independent Research**
smit.anand033@gmail.com

**ABSTRACT**

Can changing which matrices receive 2, 4, or 8 bits improve a small language model without increasing its parameter storage? I compare two parent-selection rules in the same genetic algorithm: a binary-tournament variant and an Alternis-inspired rule. A separate baseline samples repaired random allocations without crossover or mutation. All three use the same MLX affine quantizer, byte ceiling, initialization and candidate allowance within each study. An exploratory study on Qwen3.5-2B evaluates 90 allocations per run. The main study on Qwen3-1.7B evaluates 500 distinct allocations per run over 10 paired seeds. Against uniform four-bit, tournament reduces held-out next-token negative log-likelihood (NLL) by 0.0616 on AG News and 0.0152 on IMDb on average, improving all 10 seeds on each corpus at identical parameter storage. Alternis improves five seeds and ties five on AG News; on IMDb its mean gain is near zero. Random returns uniform four-bit in every seed. The tournament representative answers 156 of 400 ARC-Challenge questions correctly versus 163 for uniform; no accuracy gain is demonstrated on this subset. The result favors this tournament variant for text prediction under the tested protocol, not for compression ratio or task accuracy.

## 1 Introduction

Quantizing a language model means storing its weights with fewer bits. This can make the difference between a model that fits on a device and one that does not. The usual starting point is to give every weight matrix the same precision. But a bit saved in one matrix need not cost the model as much as a bit saved in another. Mixed-precision quantization takes that difference seriously: keep more precision where rounding changes the model's predictions most, and recover the space from matrices that tolerate coarser weights [26,27].

The difficulty is deciding where those bits belong. Suppose one allocation preserves an attention matrix and another uses those bits in a feed-forward matrix instead. Each makes a different compromise. Since the output of one part of the model feeds into another, a choice that works well in one allocation may not work as well in the other. What matters is how the choices work together, not just which matrix looks most sensitive on its own.

This is where the idea connects to earlier work my coauthors and I did on genetic algorithms. A genetic search builds a new candidate by combining parts of existing candidates. In a tournament, the better candidate in a small contest is chosen as a parent [34]. In our earlier method, Alternis, higher- and lower-ranked candidates are placed next to one another before parents are chosen from a neighborhood [1]. The intuition is that a candidate can rank poorly overall and still contain one choice worth passing on.

For quantization, that choice might be to preserve a matrix that the strongest candidates have all rounded too aggressively. Keeping a different allocation in the parent pool gives that choice a chance to survive. The opposite is just as plausible: the strongest allocations may already contain combinations that work well together, and mixing in a weaker parent may break them apart. The question behind this paper is whether a search makes better use of a fixed supply of bits by favoring the best allocations it has found, or by giving less successful ones a chance to contribute.

## 2 Background and Terminology

### 2.1 Small language models and transformer matrices

An autoregressive language model predicts the next token from preceding tokens. Transformer blocks repeatedly apply linear maps for attention and feed-forward computation. Each linear map stores a weight matrix. A model with billions of parameters may require several gigabytes even when each parameter uses only two bytes. In this paper, 'SLM' describes the tested 1.7B- and 2B-parameter class; it does not imply a universal size threshold.

Matrices are a useful precision-allocation unit because they align with model modules and quantized linear kernels. The allocation does not remove matrices or weights. It changes how many bits represent the weights of an eligible matrix. Fixed and ineligible parameters follow the common base recipe and are included in total byte accounting.

### 2.2 Affine group quantization

MLX affine quantization partitions a matrix into groups of 64 weights. For group g and within-group position j, $q_{g,j}$ is the stored quantized integer, $s_g$ is the group scale, $a_g$ is the affine bias, and $w_{g,j}^{recon}$ is the reconstructed weight.

$$w_{g,j}^{recon} = s_g \, q_{g,j} + a_g \tag{1}$$

Fewer bits reduce packed-value storage but increase rounding and clipping error. Smaller groups can adapt more locally but require more metadata; group size is fixed at 64 in both studies.

The experiments use weight-only post-training quantization (PTQ). The pretrained weights are converted after training; no gradient-based model retraining is performed. Activations and the KV cache are outside the searched weight allocation. During inference, quantized weights are consumed by quantized linear operations. Weight storage, runtime memory, activation quantization, and KV-cache quantization are separate quantities and should not be conflated.

Equation (1) defines reconstruction, not the complete conversion rule. Each eligible matrix is passed to MLX-LM/MLX's native to_quantized(group_size=64, bits=b) conversion with affine mode, grouping consecutive weights along the last dimension. The stored integer range is 0 through $2^b$-1, packed in uint32 arrays; scales and affine biases use the conversion's floating dtype inherited from the source weights. Native scale selection and integer rounding are those of the pinned MLX implementation [41,42] (mlx.core.quantize; https://ml-explore.github.io/mlx/build/html/python/_autosummary/mlx.core.quantize.html), with no separately tuned clipping rule. Layer biases and other fixed arrays are counted separately. Appendix H and the accompanying source archive specify the executable conversion and exact per-matrix costs.

### 2.3 Metrics

For an equal-length token block D, T is the number of predicted tokens, $x_t$ is token t, $x_{<t}$ is its preceding context, b is the matrix-bit genome, and $p_b$ is the probability assigned by the quantized model.

$$L_D(b) = -(1/T)\ \Sigma_{t=1}^{T} \log p_b(x_t \mid x_{<t}) \qquad (2)$$

Lower NLL is better. Perplexity is exp(NLL), also lower-is-better but nonlinear. ARC accuracy is the fraction of fixed multiple-choice questions answered correctly using length-normalized answer likelihoods. NLL and ARC accuracy measure different behaviors; improvement in one does not guarantee improvement in the other.

### 2.4 Deployment and search controls

Uniform four-bit quantization is the no-search deployment control. Random sampling is the attribution control for search: it explores the same feasible space with the same allowance but proposes candidates independently. Tournament and Alternis are parent-selection policies layered on the same quantizer. An improvement over uniform but not random would support generic search; an improvement over random and another genetic selector is evidence about search policy under the frozen protocol.

The complete experimental sequence and the boundary between search, validation, and held-out evaluation are summarized in Figure 1.

## 3 Related Work

### 3.1 Foundations and uniform PTQ

Integer-only quantization established practical affine scaling and zero-point style deployment for neural networks [2]. Modern LLM PTQ targets much larger transformer matrices and more aggressive precision. GPTQ minimizes layer reconstruction error with approximate second-order information [3]. AWQ uses activation statistics to protect salient channels [4]. SmoothQuant transfers activation difficulty into weights [5], while OmniQuant learns clipping and equivalent transformations from calibration data [6]. These approaches change the candidate quantizer itself rather than the policy that allocates matrix bit-widths.

### 3.2 Outliers, rotations, rounding, and codebooks

LLM.int8() isolates activation outliers [7]; OWQ protects outlier-sensitive weights [8]; SpQR and SqueezeLLM combine dense low-bit storage with sparse exceptions [9,10]. AdaRound adapts weight rounding to data and task loss [22]; BRECQ reconstructs blocks [21], and AutoRound optimizes rounding with signed gradients [13]. RPTQ reorders channels [14]. QuaRot and SpinQuant apply rotations to suppress outliers [15,16]. QuIP and QuIP# combine incoherence processing with low-bit representations [11,17], while QTIP, GPTVQ, VPTQ, and AQLM exploit trellises, vectors, or additive codebooks [12,18–20]. These papers shape the landscape a search policy would see, but they do not answer the isolated selector comparison studied here.

### 3.3 Automated mixed-precision allocation

HAQ and AutoQ learn per-layer or per-kernel policies with reinforcement learning [24,25]. HAWQ and HAWQ-V2 use Hessian information to rank layer sensitivity and constrain the mixed-precision space [26,27]. FracBits relaxes discrete bit choices into fractional values [28]. APQ jointly searches architecture, pruning, and quantization with an accuracy predictor and evolutionary search [29]. EMQ uses diversity-promoting selection to evolve a scoring formula that ranks mixed-precision configurations [30]. Here, selection acts on the bit allocations themselves, not on a formula for scoring them. LLM-MQ uses layer sensitivity and integer programming to assign bit-widths under a weight-memory budget [23]. MixLLM assigns precision across output features [38]. We instead assign one width per matrix and test how the search chooses parents.

### 3.4 Evolutionary and population-based compression

EvoPress is a close precedent: it uses evolutionary search for nonuniform LLM compression [31]. UDP uses a genetic search to choose which layers receive higher precision under a fixed budget [47]. Our question is how the choice of parents affects the allocations found under the same search budget. Miccini et al. apply genetic algorithms to operator-level sub-8-bit mixed precision for recurrent networks [32]. CLADO explicitly models cross-layer quantization dependencies [36], and MetaMix learns a precision searcher for activation quantization [37]. These studies establish that population search and cross-layer allocation are viable, but none provides the same equal-candidate comparison of Alternis-inspired selection, binary tournament selection, and random sampling on an MLX SLM.

### 3.5 Search-policy foundations

Random search is a necessary equal-budget baseline because sophisticated search can appear effective merely by evaluating more candidates [33]. Tournament selection provides tunable selection pressure and has known behavior under noisy fitness [34]. NSGA-II [35] is relevant when quality, size, speed, and preparation cost form a true multiobjective problem. The present study uses one hard size constraint and minimizes NLL, so it does not implement a Pareto algorithm.

| Family | Representative work | What changes | Directly tests selector? |
|---|---|---|---|
| Second-order / saliency | GPTQ, AWQ, HAWQ | Quantizer or sensitivity | No |
| Rotation / rounding | QuaRot, SpinQuant, AutoRound | Weight transformation | No |
| Codebook / vector | AQLM, QuIP#, GPTVQ | Representation | No |

| Family | Representative work | What changes | Directly tests selector? |
|---|---|---|---|
| Automated allocation | HAQ, AutoQ, FracBits | Bit policy | Adjacent |
| Evolutionary compression | APQ, EMQ, EvoPress, UDP | Predictor/proxy/policy | Closest |
| This work | Two GAs + random control | Parents (GAs only) | GA pair only |

**Table 1.** Related-work taxonomy. ‘Closest’ does not mean protocol-equivalent; most prior work changes more than the selector.

# 4 Problem Formulation

## 4.1 Genome and exact byte constraint

Let $W_1,...,W_L$ be the L eligible weight matrices. A genome $b=(b_1,...,b_L)$ assigns bit-width $b_i$ in {2,4,8} to matrix $W_i$. Let F denote bytes occupied by fixed or ineligible parameters; $Q_i$, $S_i$, and $A_i$ denote the packed integer values, group scales, and affine biases for matrix i. The argument $b_i$ indicates that these tensors are produced using that matrix's assigned bit-width.

$$C(b) = F + \Sigma_{i=1}^{L} [\text{bytes}(Q_i(b_i)) + \text{bytes}(S_i(b_i)) + \text{bytes}(A_i(b_i))] \quad (3)$$

The reference budget is $B_4=C((4,...,4))$, the exact measured cost of assigning four bits to every eligible matrix.

$$\min_b L_{search}(b) \quad \text{subject to} \quad C(b) \le B_4 \quad (4)$$

The genome b specifies one width per eligible matrix. Feasibility requires $C(b) \le B_4$: the constraint is a ceiling, not an instruction to use every byte. In the recorded V3 searches, all 15,000 entries reached the same 968,001,536 parameter bytes as uniform four-bit. Packed weights, scales, affine biases and fixed parameters count; tokenizer files, configuration, binaries, activations and KV cache do not.

## 4.2 Worked allocation example

| Matrix | M1 | M2 | M3 | M4 | M5 | M6 | Total units |
|---|---|---|---|---|---|---|---|
| Uniform | 4 | 4 | 4 | 4 | 4 | 4 | 24 |
| Feasible mixed | 2 | 2 | 4 | 4 | 4 | 8 | 24 |
| Infeasible child | 8 | 8 | 4 | 4 | 4 | 8 | 36 |

**Table 2.** Toy equal-shape example in bit-units. Real matrices have different shapes and metadata, so the implementation measures bytes rather than summing nominal bits.

The feasible mixed allocation protects M6 with 8 bits by assigning M1 and M2 only 2 bits. A crossover can produce the infeasible child. Repair stochastically demotes 8 to 4 to 2 bits until measured bytes fit, then promotes randomly selected genes while a single upgrade still fits. Repair is part of the frozen search landscape: it influences which feasible genomes are reachable and must be shared across selectors.

## 4.3 Sensitivity-informed initialization

A calibration subset, also contained in the search data, evaluates each matrix at 2 and 8 bits relative to an all-4 reference. For matrix i and trial precision r in {2,8}, the single-matrix loss change is

$$\Delta_{i,r} = L_{sens}(4,...,r_i,...,4) - L_{sens}(4,...,4) \quad (5)$$

Here $L_{sens}$ is NLL on the sensitivity-probe tokens. A positive value means the trial precision worsens NLL; a negative value means it improves NLL. An additive proxy starts from all-2 and greedily purchases the upgrade with the largest predicted loss reduction per added byte. This allocation and perturbations of it join uniform and random feasible genomes in the initial population. The proxy is only an initializer; final candidate fitness is measured NLL.

# 5 Search Algorithms

## 5.1 Shared population mechanics

Within a study, all arms share the same model, eligible matrices, quantizer, byte constraint, data, seed set, initial population, candidate allowance, duplicate rejection, finalist pool, and validation rule. The two evolutionary arms also share elitism, crossover, mutation, and repair. The random arm has no parents and therefore no crossover; it proposes fresh random repaired genomes while retaining the best observed genome only for reporting and generation structure.

Crossover samples two cut positions s<e and transplants one contiguous segment from parent b into parent a:

$$\text{child} = a[:s] + b[s:e] + a[e:] \quad (6)$$

Each gene mutates with probability 1/L by choosing one of the other two bit-widths. A global seen set rejects duplicate genomes, so the allowance counts distinct scored allocations rather than proposal attempts.

V2 evaluates 30 initial candidates, 29 new candidates in each of two complete offspring generations, then two new candidates in a final partial generation: 30+29+29+2=90. V3 evaluates 18 initial candidates and 30 complete offspring generations: 16 new candidates in 28 generations and 17 in generations 15 and 30, for 18+28(16)+2(17)=500. Each offspring generation also retains one already-scored elite. Thus V3 populations usually contain 17 individuals, with 18 at initialization and generations 15 and 30. Candidate allowance excludes rejected duplicate proposals and reuse of elite scores.

## 5.2 Binary tournament

A tournament parent index is the lower-NLL member of two population members sampled without replacement, with deterministic index tie-breaking. Three independent tournaments fill three pool positions once per generation. Winners can repeat across positions. Each offspring samples two distinct pool positions without replacement; these positions can refer to the same individual, so self-mating is possible. The pool is reused for all offspring in that generation. This describes the implemented tournament variant, not every binary-tournament GA.

## 5.3 Alternis adaptation

Original Alternis arranges chromosomes in alternating fitness order and chooses a contiguous neighborhood [1]. Our minimization adaptation sorts population indices by (loss,index), then constructs [worst,best,second-worst,second-best,...]. One valid interior center is drawn uniformly per generation. The center and its immediate neighbors form three distinct individual indices. Each offspring samples two distinct positions from this fixed pool. Parents may be reused across offspring. Unlike the tournament pool, this pool has no repeated individuals. The comparison includes each selector's implemented parent-reuse behavior. It is an Alternis-inspired operator, not an unchanged reproduction of the original application.

### 5.4 Elitist random sampling

The random control proposes bit-widths from a frozen study-specific categorical prior, applies exact-byte repair, rejects duplicates, and scores the result. Both V2 and V3 use proposal weights (.15,.80,.05) for (2,4,8), as verified against the frozen InformedSpace.random source. These priors are shared by the corresponding random components and frozen before evaluation. The current best is retained, but it does not generate proposals; the arm is not a genetic algorithm.

### 5.5 End-to-end pseudocode

```
P = SharedInit(seed); H = empty_set; A = archive
for x in P: score x; add x to H and A
checkpoint initial population (generation 0)
for g=1,...,G:
  P_next = {argmin_{x in P} L_search(x)}
  R = SelectParents(P,method)   # absent for random
  while scheduled offspring remain:
    y = Repair(RandomGenome()) if random
        else Repair(Mutate(Crossover(parent(R),parent(R))))
     if y not in H: score y; checkpoint y; add y to H, A and
P_next
  P = P_next; checkpoint generation statistics
Q = unique(top-3(A) + {uniform-4, informed})
x* = argmin_{x in Q} L_validation(x); freeze x*
only then evaluate held-out text, task, benchmarks, and export
```

## 6 Experimental Design

| Setting | V2 exploratory | V3 controlled follow-up |
|---|---|---|
| Model | Qwen3.5-2B | Qwen3-1.7B |
| Eligible genes | 186 | 196 |
| Bits / group | {2,4,8} / 64 | {2,4,8} / 64 |
| Population | 30 initial; 30,30,3 | 18 initial; usually 17 |
| Candidates / arm / seed | 90 | 500 |
| Offspring generations | 2 full + 1 partial | 30 complete |
| Paired seeds | 10 | 10 |
| Candidate entries across runs | 2,700 | 15,000 |
| Sequence length | 512 | 256 |
| Search tokens | 16,384 | 8,192 |
| Validation tokens | 65,536 | 32,768 |
| Held-out text | 131,072 | 65,536 fresh |
| Task | 400 ARC-Easy | 400 ARC-Challenge |
| Uniform-4 bytes | 1,059,320,512 | 968,001,536 |

**Table 3.** Frozen protocols. V2 and V3 differ in model, data, population, allowance, sequence length, and task; their numeric outcomes must not be pooled or treated as a causal depth ablation.

### 6.1 Data separation

V2 search uses revision-pinned WikiText-2 [43] and AG News [44] training text. AG News training records are shuffled with a frozen seed: 0–5,000 support calibration and 5,000–10,000 validation. The official AG News test split is held out. Four hundred ARC-Easy test questions [46] exclude 64 identifiers used earlier.

V3 uses frozen, disjoint AG News [44] and IMDb [45] partitions. Calibration interleaves 64 news and 64 IMDb blocks; search takes the first 32 interleaved blocks (8,192 tokens), while validation contains 32,768 separate tokens. Corrected wide30 held-out evaluation uses AG News training-split shuffled indices 25,000–30,000 and IMDb official-test shuffled indices 5,000–10,000, contributing 32,768 tokens per corpus. Four hundred ARC-Challenge items [46] exclude every earlier v1/v2/v3 identifier.

### 6.2 Two levels of validation selection

For each method-seed run in both studies, the three lowest search-NLL genomes from the complete scored archive plus shared informed and uniform candidates form a deduplicated validation shortlist. Minimum validation NLL selects the finalist, with lexicographic genome tie-breaking. Seed-level text comparisons use all 10 finalists. V2 evaluates ARC-Easy and timing for each finalist; Table 4 reports mean seed accuracy, with identical genomes sharing cached evaluations. V3 additionally selects one representative per method by (validation NLL, genome identifier) for ARC-Challenge, timing and export. These single-representative V3 measurements are not 10-seed task or timing inference.

### 6.3 Checkpoint and integrity controls

NLL evaluations are cached in SQLite by split and genome identifier; ordered per-run candidate histories are saved in JSON checkpoints. Resume deterministically replays random-number generation and rejects genome or generation mismatches. A global seen set guarantees distinct candidate counts. Generation files record population size, total evaluations, best search NLL, and mean pairwise Hamming distance. Source and data SHA-256 manifests are frozen and checked. Completion requires all expected scores and generation records, byte-budget compliance, held-out completeness, and a selected-export reload with bounded logit difference.

### 6.4 Statistical scope

For paired seeds, positive NLL reduction favors the searched method, as defined formally in Equation (8). V2 uses the stored 10,000-resample paired bootstrap. V3 reports paired-t intervals over the 10 stochastic search seeds, with exact sign tests as a sensitivity check. These quantify search randomness conditional on one model and fixed corpora. Multiple comparisons are unadjusted and exploratory; they do not establish generalization across models, datasets, or deployments.

## 7 V2 Exploratory Study

### 7.1 Selection-policy result

| Method | Mean test NLL (lower) | PPL (lower) | ARC-Easy (higher) | NLL wins/ties |
|---|---|---|---|---|
| Uniform-4 | 3.62258 | 37.434 | 60.25% | -- |
| Random | 3.62258 | 37.434 | 60.25% | 0 / 10 |
| Tournament | 3.62119 | 37.382 | 59.68% | 2 / 8 |
| Alternis | 3.62258 | 37.434 | 60.25% | 0 / 10 |

**Table 4.** V2 held-out outcomes at the uniform-four budget. Wins/ties are versus uniform for NLL.

Random and Alternis select the uniform allocation for all 10 seeds and therefore tie exactly. Tournament improves two seeds and ties eight, yielding a mean NLL reduction of 0.001384. Alternis minus tournament is +0.001384 NLL with bootstrap interval [0,+0.003330], so Alternis does not beat tournament. Tournament's small NLL gain does not transfer to ARC-Easy: its mean accuracy is 59.68% versus 60.25% for uniform.

### 7.2 Quantizer baselines

Figure 2 places the v2 uniform, sensitivity, GPTQ, and combined baselines on the same storage-versus-metric axes.

At 1.059 GB, GPTQ-4 lowers test NLL from 3.62258 to 3.57643 and raises ARC-Easy from 60.25% to 62.75%. Alternis-allocation-plus-GPTQ is exactly equal to GPTQ-4 because validation selected the all-4 allocation; it supplies no synergy evidence. Uniform 3-bit is smaller but degrades sharply, while uniform 6- and 8-bit approach the higher-precision model at increased size.

### 7.3 Separate AWQ compatibility control

A Qwen3-1.7B control compares higher precision, uniform MLX-4, GPTQ-4, and AWQ-4 on that model only. AWQ has the best 4-bit NLL (3.70452) versus uniform MLX (3.72651) and GPTQ (3.74152), while GPTQ and AWQ improve ARC accuracy. Because this control uses a different model from the v2 main experiment and does not pair AWQ with the allocation search, it cannot rank AWQ against main-model Alternis or tournament outcomes.

## 8 V3 Controlled Follow-up

### 8.1 Validation

Uniform four-bit is always on the validation shortlist. Choosing the lowest validation NLL therefore cannot yield a value worse than uniform on those same tokens. A strict win means that the shortlist contained a better choice on validation; held-out text tests whether the gain transfers. Uniform scores 3.75610. Random selects it in all 10 seeds. Tournament averages 3.72122 (gain 0.03489; paired 95% interval [0.02407,0.04570]), improving all 10 seeds. Alternis averages 3.75068 (gain 0.00543; interval [-0.00094,0.01180]), improving five. Tournament beats Alternis in nine seeds, by 0.02946 NLL on average.

### 8.2 Corrected fresh held-out text

The original evaluation read earlier token arrays. Without changing the model choices, we rescored the frozen finalists on the intended held-out text: withheld AG News training-split records (not its official test set) and held-out IMDb official-test records. Table 5 and Figure 3 report only the corrected scores.

| Corpus / method | NLL (lower) | Gain (higher) | 95% gain interval | W/T/L |
|---|---|---|---|---|
| AG / Random | 3.85462 | 0 | [0,0] | 0/10/0 |
| AG / Alternis | 3.84491 | .00970 | [.0007,.0188] | 5/5/0 |
| AG / Tournament | 3.79299 | .06163 | [.0465,.0768] | 10/0/0 |
| IMDb / Random | 3.74209 | 0 | [0,0] | 0/10/0 |
| IMDb / Alternis | 3.74170 | .00038 | [-.0061,.0069] | 3/5/2 |
| IMDb / Tournament | 3.72689 | .01520 | [.0071,.0233] | 10/0/0 |

**Table 5.** NLL on corrected held-out text. W/T/L counts paired seeds with lower, equal or higher NLL than uniform four-bit on the same corpus.

Tournament improves all 10 seeds on each corpus. A two-sided exact sign test for 10 wins in 10 trials gives p=0.00195 per corpus before multiplicity adjustment. Relative to uniform, Alternis records 5 wins, 5 ties and no losses on AG News, versus 3 wins, 5 ties and 2 losses on IMDb. Tournament also beats Alternis on all 10 AG seeds and eight of 10 IMDb seeds. These are exploratory sensitivity checks, not model-family inference.

### 8.3 Allocation and equal-byte meaning

The validation-selected tournament representative assigns 11 matrices to 2 bits, 170 to 4 bits, and 15 to 8 bits. The Alternis representative assigns 9, 183, and 4, respectively. Both occupy exactly 968,001,536 parameter bytes, 28.13% of the higher-precision parameter bytes. Tournament moves more matrices away from four bits. These counts do not say which individual changes caused its NLL gain.

### 8.4 Downstream task result

On the 400-item ARC-Challenge subset, the tournament representative answers 156 questions correctly (39.00%), versus 163 for uniform and random (40.75%) and 158 for Alternis (39.50%). Relative to uniform, tournament gains 18 answers and loses 25 (two-sided exact paired McNemar p=0.36). We found no evidence of an accuracy gain on this subset. One validation-selected representative per method is not a test of reasoning accuracy across seeds or tasks.

### 8.5 Inference and export

Generation throughput was similar across representatives in three local trials after warm-up. Without a documented chip and memory configuration or more trials, these measurements do not establish a speed ranking. Parameter bytes and measured peak MLX allocation were equal across the benchmarked representatives. The selected tournament export reloaded with zero maximum logit difference on the input described in Appendix H.4.

## 9 Search Dynamics

Figure 4 shows how best search NLL and population diversity evolve across the 30 v3 generations.

All arms begin with mean best search NLL 3.773886 and mean pairwise Hamming distance 70.77. Random sampling remains at the same best NLL through generation 30 while sustaining high diversity (69.87). Tournament steadily improves to 3.738768 and contracts diversity to 14.77. Alternis improves only to 3.768204 while retaining intermediate diversity (30.16).

The traces are consistent with greater exploitation by this tournament variant: lower population diversity accompanies lower search NLL and better validation and held-out text NLL. Alternis retains more population diversity, but this does not yield equally strong allocations. These are descriptive associations, not a causal isolation of selection pressure. Population Hamming distance does not directly measure parent-pair diversity or useful diversity. A mechanism claim would require parent-pair logging and matched operator ablations.

This does not prove tournament is universally optimal. A different initializer, mutation rate, repair operator, budget, or multimodal objective could make premature convergence harmful and increase the value of Alternis. The result is conditional on the complete search system, not an intrinsic ranking independent of context.

For population P of size m, including the retained elite, the recorded diversity is the mean count of differing bit assignments over all unordered pairs:

$$D(P) = [2 / (m(m-1))] \Sigma_{i<j} \Sigma_{k=1}^{L} 1[b_{i,k} \neq b_{j,k}] \quad (7)$$

The indicator is one when the two genes differ and zero otherwise. D is unnormalized, ranges from 0 to L, and is computed on each population before averaging across 10 seeds in Figure 4. The elite is included; this metric does not describe the sampled mating pool.

# 10 Statistical Interpretation

## 10.1 Inferential unit and uncertainty

Each seed changes stochastic initialization perturbations, parent samples, crossover cuts, mutations, and repair choices. The paired seed is therefore the experimental unit for selector comparisons. The data corpora and model are fixed. AG News and IMDb reuse the same selected allocations and are two evaluation corpora, not independent replications of the search experiment. A narrow confidence interval indicates stable search behavior for this frozen setting; it does not imply stability across another model family or task.

For a method comparison, $d_k$ is the metric difference for paired seed k. For held-out NLL, a positive reduction favors the method:

$$d_k = NLL_{control,k} - NLL_{method,k} \quad (8)$$

Let mean(d) be the mean of the n=10 paired differences and $s_d$ their sample standard deviation. Unless a table explicitly says bootstrap, the reported two-sided interval is

$$\text{mean}(d) \pm t_{0.975,n-1}\, s_d / \text{sqrt}(n) \quad (9)$$

where t is the Student-t critical value. Accuracy differences use percentage points.

## 10.2 Multiple comparisons

The manuscript reports three policies, validation, two held-out text corpora, and ARC outcomes. No multiplicity correction was preregistered. Confidence intervals and sign tests should therefore be read as descriptive and exploratory. The strongest evidence is the consistent direction of tournament gains across all paired seeds and both fresh text corpora, combined with a null downstream task result.

## 10.3 Selection layers and winner's curse

Seed-level text statistics average 10 validation-selected finalists. In V3, task, speed and export use an additional validation-selected method representative. Selecting the best seed can make its validation value optimistic; the paired text claim instead uses all seed finalists. V2 task results are means over seed finalists and must not be interpreted as V3-style single-representative measurements.

## 10.4 What counts as a scientific win

A selector ‘wins’ only relative to a named control and metric. Tournament lowers held-out text NLL relative to uniform, random, and Alternis at equal parameter bytes and the same candidate allowance. It does not win on ARC accuracy, storage, or demonstrated inference speed. Alternis's negative result is still informative because it shows that greater diversity is insufficient under this search landscape.

# 11 Comparison with Prior Studies

| Work | Model domain | Allocation unit | Search mechanism | Relation |
|---|---|---|---|---|
| HAQ [24] | vision | layer | reinforcement learning | policy precedent |
| HAWQ-V2 [27] | vision | layer | Hessian + Pareto | sensitivity precedent |
| APQ [29] | vision | architecture/prune/bits | predictor + evolution | evolution precedent |
| EMQ [30] | vision | proxy formula | genetic programming | evolves proxy |
| Miccini et al. [32] | GRU | operator | genetic algorithm | direct mixed precision |
| EvoPress [31] | LLM | compression policy | evolutionary search | LLM precedent |
| UDP [47] | LLM | layers to promote | genetic search | budgeted allocation |
| This work | Qwen SLM | matrix | 2 GA rules + random | GA pair isolates selector |

**Table 6.** Closest automated-allocation studies. Protocols, models, and quantizers differ; the table maps methodological roles rather than claiming leaderboard comparability.

EvoPress searches nonuniform LLM compression policies [31]; UDP searches which layers to promote to higher precision [47]. Our contribution is a controlled comparison of two implemented parent selectors within a shared matrix-bit GA, plus an equal-candidate random-proposal control. Neither EvoPress nor UDP is a head-to-head experimental baseline here. The comparison fixes the MLX affine evaluator, 2/4/8-bit matrix space, byte ceiling, initialization, repair, mutation, crossover, data and candidate allowance for the two genetic arms. Random removes parent selection, crossover and mutation together; its comparison with a GA measures the combined proposal mechanism, not selection alone. The conventional uniform model incurs no search cost.

The quantizer literature is complementary. GPTQ or AWQ could define a different candidate evaluator; QuaRot or SpinQuant could change the error landscape; AQLM or GPTVQ could change the representation entirely. Testing the same selectors on top of compatible backends is a future factorial study. Current v2 shows why this matters: GPTQ produces a larger equal-byte main-model gain than allocation search, while Alternis+GPTQ collapses to ordinary GPTQ because the selected allocation is uniform.

# 12 Practical Significance

Uniform four-bit remains the practical baseline for users who need immediate conversion. It requires no search and produced the same storage and nearly the same measured inference behavior as searched representatives. Tournament search is a one-time preparation process that may be justified when a model will be deployed repeatedly and a modest NLL improvement has operational value.

The practical gain in V3 is better next-token prediction on the two held-out corpora at the same parameter storage as uniform four-bit. It is not a reduction in storage.

A deployment decision should include the target workload. If the device primarily performs language modeling or domain text scoring similar to the tested corpora, the NLL gain may matter. No benefit was observed on the tested 400-item ARC-Challenge subset using the global validation-selected representatives; this single evaluation cannot establish performance on reasoning or multiple-choice tasks generally. If preparation time is constrained, uniform conversion or a strong one-shot quantizer may dominate.

## 13 Limitations and Future Work

- Only two Qwen-family models are complete, and v2/v3 use different protocols; model-family generalization is untested.
- The controlled v3 selector result uses one affine weight-only MLX backend, group size 64, bit set {2,4,8}, and one exact byte target.
- The primary evidence is token NLL on two fixed corpora; the single 400-item ARC task does not cover instruction following, safety, factuality, coding, or long context.
- Seed intervals characterize stochastic search, not a random sample of models or datasets, and reported comparisons are unadjusted for multiplicity.
- Repair, initialization, and candidate encoding shape the landscape; another implementation could change the selector ranking.
- Search cost is substantial and was not normalized against all modern quantizer preparation costs under a common system protocol.

High-value follow-ups include repeating the exact selector comparison across model families and byte budgets; using nested or cross-domain validation; testing GPTQ-, AWQ-, HQQ-, AutoRound-, and rotation-based compatible candidate evaluators; adding NSGA-II or Bayesian optimization for genuine size/quality/preparation tradeoffs; and performing ablations that independently vary generations, population size, candidate allowance, initialization, mutation, and repair.

Alternis may deserve targeted testing where diversity is plausibly valuable: deceptive landscapes, multiple local optima, multiobjective frontiers, heterogeneous backend constraints, or intentionally weak initialization. Those experiments should be predeclared and should compare equal numbers of unique candidate evaluations.

## 14 Conclusion

At the uniform-four byte ceiling, the exploratory Qwen3.5-2B study found little separation between the selectors. In the separate Qwen3-1.7B study, this tournament variant had lower mean held-out NLL than uniform, repaired random sampling and Alternis on both corpora, without increasing parameter storage. Its populations also became less diverse; the study does not establish that diversity caused the NLL difference. The 400-item ARC-Challenge result showed no demonstrated accuracy gain, and the timing trials did not establish a speed advantage.

# Figures  Full-Size Experimental Figures

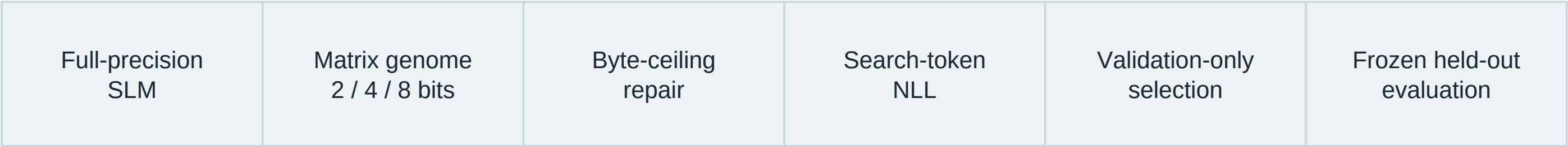

**Figure 1.** Experimental pipeline. Read left to right. The two genetic arms change the parent selector; random uses fresh repaired proposals without crossover or mutation. The model, quantizer, byte ceiling, data and evaluation protocol are shared.

A genome assigns 2, 4, or 8 bits to each eligible matrix. Search-token NLL provides fitness, validation freezes the finalist, and held-out text and ARC are evaluated only afterward.

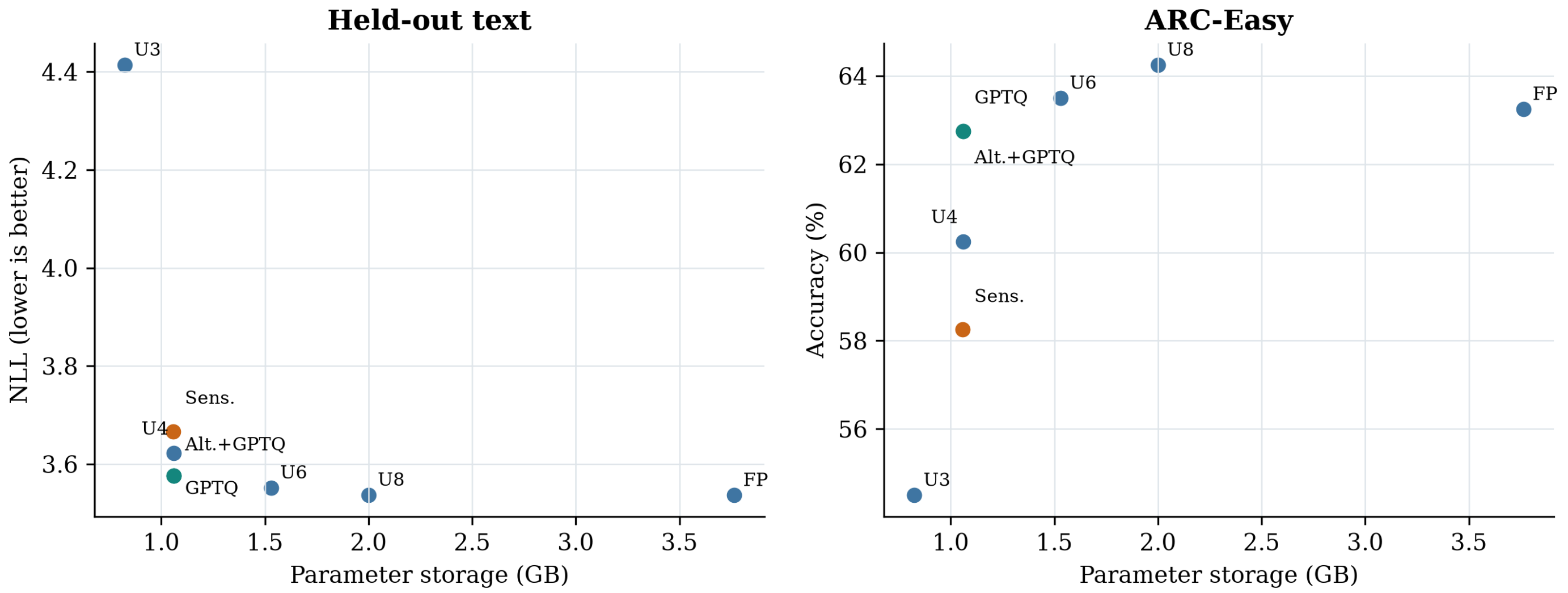

**Figure 2.** V2 Qwen3.5-2B storage and metrics. Lower held-out NLL is better; higher ARC-Easy accuracy is better. U3/U4/U6/U8 are uniform bit-widths, FP is higher precision, Sens. is sensitivity allocation, and Alt.+GPTQ is the combined allocation-plus-GPTQ control.

Points with similar measured parameter storage can be compared directly. GPTQ changes the quantizer itself, so its gain cannot be attributed only to the search selector.

**Reading guide for Figures 1–2.** Figure 1 separates the three stages that can influence a result: search proposes and scores candidate genomes, validation chooses among the frozen shortlist, and held-out evaluation estimates transfer after selection. Figure 2 then shows why storage alone is insufficient: models at nearly the same measured bytes can have different NLL and ARC accuracy, while the best point can depend on which metric is prioritized.

## Figures Full-Size V3 Results

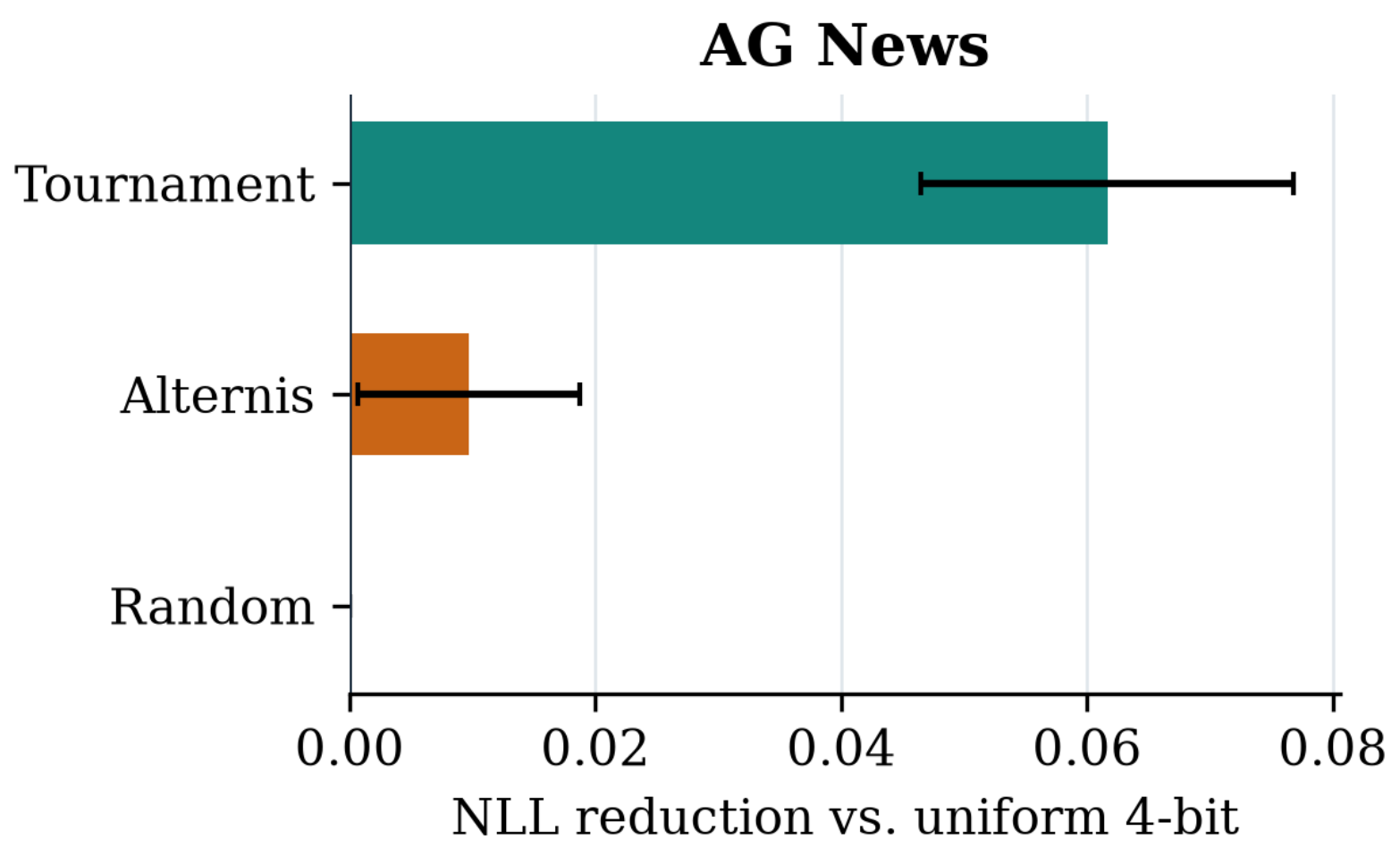

**Figure 3.** Corrected fresh held-out NLL reduction at equal parameter bytes. Positive values favor the searched allocation; zero is a tie. Bars are means across 10 paired seeds and whiskers are paired 95% t intervals. Panel x-axis scales differ; compare the labeled NLL units, not bar lengths across panels.

The intervals describe stochastic search repeatability for the fixed model and corpora, not uncertainty across different model families or datasets.

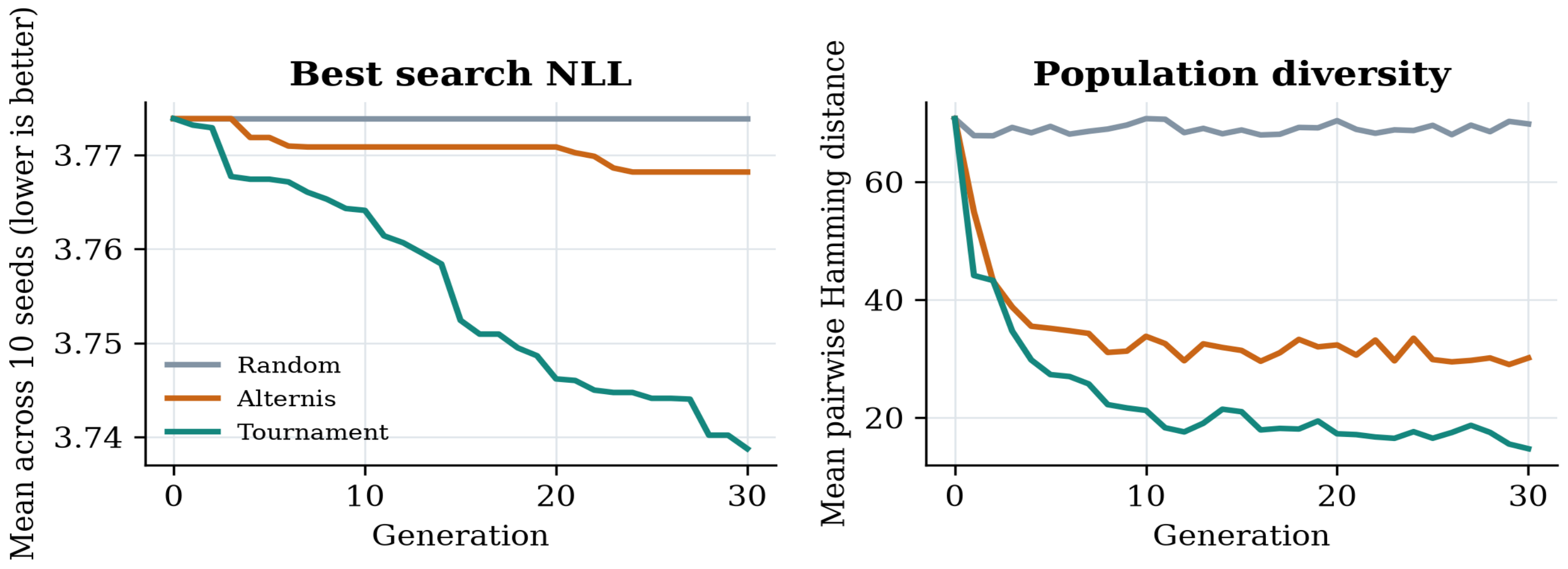

**Figure 4.** V3 search dynamics across 30 generations. Lower best-search NLL is better. Higher mean pairwise Hamming distance means that more matrix bit assignments differ within the population.

These curves use the search objective and explain optimizer behavior; they are not held-out evidence.

## A Appendix: Per-Seed V3 Evidence

| Seed | Val T | Val A | AG T | AG A | IMDb T | IMDb A |
|---|---|---|---|---|---|---|
| 101 | 3.69996 | 3.75610 | 3.78275 | 3.85462 | 3.70782 | 3.74209 |
| 211 | 3.71361 | 3.75610 | 3.77140 | 3.85462 | 3.73197 | 3.74209 |
| 307 | 3.73069 | 3.75097 | 3.78634 | 3.84819 | 3.73858 | 3.73593 |
| 401 | 3.71077 | 3.75094 | 3.78135 | 3.83597 | 3.72487 | 3.75487 |
| 503 | 3.72995 | 3.74012 | 3.81016 | 3.82413 | 3.72690 | 3.73945 |
| 601 | 3.73594 | 3.72969 | 3.80988 | 3.82381 | 3.73869 | 3.72238 |
| 701 | 3.74336 | 3.75610 | 3.82971 | 3.85462 | 3.74008 | 3.74209 |
| 809 | 3.72371 | 3.75453 | 3.80350 | 3.84395 | 3.72916 | 3.75395 |
| 907 | 3.72610 | 3.75610 | 3.79695 | 3.85462 | 3.71942 | 3.74209 |
| 1009 | 3.69809 | 3.75610 | 3.75784 | 3.85462 | 3.71140 | 3.74209 |

**Table A1.** V3 tournament (T) and Alternis (A) NLL by seed. Uniform validation is 3.75610; uniform AG/IMDb are 3.85462/3.74209. Full-precision values, not rounded display values, drive statistics.

# B Appendix: Audit and Correction

The first wide30 text-evaluation procedure accidentally read earlier v3 arrays even though the study had already prepared fresh arrays. Search, validation scores, finalists, global representatives, ARC answers, byte measurements, benchmarks, and model export were unaffected. Reevaluation used the already-frozen genomes, hashed fresh arrays, and distinct cache keys. The authoritative heldout_text_fresh.json contains 62 rows: 30 finalists times two corpora plus uniform on each corpus. Historical scores are retained only as labeled secondary history.

The v3 completion audit verifies 30 searches by 500 distinct genomes, generation checkpoints 0-30, 15,000 recorded candidate entries (14,622 distinct genome IDs across runs), byte compliance for every score and representative, 62 corrected fresh text rows, four task groups, four inference benchmark groups, zero overlap with earlier ARC identifiers, matching source/data hashes, and selected-export reload with zero maximum logit difference.

# C Appendix: Reproduction Checklist

- Verify model revision, tokenizer, eligible matrix list, group size, bit choices, and exact uniform-four bytes.
- Verify source and data hashes before scoring; preserve paired seed identities and frozen random priors.
- Require distinct genome identifiers and exact per-run candidate counts; never count rejected duplicate proposals as evaluations.
- Reconstruct generation records and verify the selector, elite, crossover, mutation, and repair path deterministically.
- Build each validation shortlist from search-only results, choose by validation, and freeze choices before held-out access.
- Evaluate every paired finalist on the same fresh corpora; keep the global representative separate from seed-level inference.
- Export and reload the selected model; compare logits and verify exact parameter bytes.

# D Appendix: Annotated Literature Guide

## D.1 Closest search and allocation precedents

**EvoPress [31]** treats nonuniform LLM compression as a search problem. **UDP [47]** is a related allocation method: it profiles layer sensitivity and uses genetic search to decide which layers to promote. Both are precedents for searching under a budget, not measurements of how our parent-selection rules compare.

**APQ [29]** combines an accuracy predictor with evolutionary search over architecture, pruning, and quantization. It demonstrates how a learned surrogate can make a much larger design space searchable, but multiple design axes move together. Our study deliberately narrows the design space and pays the cost of directly scoring every unique candidate so that the selector comparison is not mediated by a learned predictor.

**EMQ [30]** uses diversity-promoting selection to search for proxy formulas that rank mixed-precision configurations. In our study a candidate is a matrix-level bit allocation, and its fitness is the measured next-token loss. The two studies use diversity at different points in the search.

**Miccini et al. [32]** use a genetic algorithm for sub-8-bit mixed-precision GRU quantization. This is direct evidence that a GA can operate on precision assignments, although recurrent operators, constraints, and target models differ. **HAQ, AutoQ, HAWQ-V2, and FracBits [24–28]** provide complementary policy-learning, sensitivity, and continuous-relaxation approaches. They motivate future head-to-head comparisons where every optimizer receives the same unique-candidate allowance.

## D.2 Quantizers that could replace the candidate evaluator

**GPTQ [3]** is a natural next backend because v2 already shows that second-order reconstruction improves the main model at uniform four-bit. A useful factorial experiment would compare uniform GPTQ, mixed GPTQ with random allocation, mixed GPTQ with tournament, and mixed GPTQ with Alternis on the same model. The v2 Alternis+GPTQ result is not this experiment: its selected allocation was uniform, so it collapses to ordinary GPTQ.

**AWQ [4]** protects activation-salient channels through scaling rather than storing a heterogeneous bit policy. The separate v2 control suggests AWQ is competitive on Qwen3-1.7B, but a proper allocation study requires a compatible same-model mixed-bit implementation. **OmniQuant, AutoRound, and BRECQ [6,13,21]** similarly change calibration or reconstruction. Each could make the allocation landscape smoother or sharper and thereby change which selector is effective.

**QuaRot, SpinQuant, QuIP#, and QTIP [15–18]** transform weights or use incoherent representations to reduce outlier-driven quantization error. If those transformations make layers more uniformly robust, matrix-level mixed precision may have less room to help. Conversely, a few remaining sensitive matrices could make allocation more valuable. This interaction is scientifically interesting because it tests whether search-policy gains survive a stronger error model.

**AQLM, GPTVQ, and VPTQ [12,19,20]** use additive or vector representations. Their cost is not captured by the affine metadata formula used here, so exact-byte repair must be redesigned rather than reused. The broader lesson is that a search genome, feasibility function, and serialization cost model form one coupled experimental object.

## D.3 Why the classical search references matter

Bergstra and Bengio [33] explain why random search is a serious baseline rather than a placeholder: when only a subset of dimensions matters, random exploration can outperform structured grids. Our random arm makes a related attribution test. Its V3 outcome favors the implemented genetic proposal mechanism over this particular repaired random prior; it does not isolate parent selection from crossover and mutation.

Miller and Goldberg [34] analyze tournament selection and noise. Candidate NLL is deterministic for a frozen model and token array, but it is still a limited proxy for validation and downstream quality. Tournament pressure can amplify candidates that exploit search-token peculiarities. The separate validation shortlist and fresh held-out evaluation are therefore essential safeguards.

NSGA-II [35] is relevant when no single hard byte target captures deployment value. A future study could jointly optimize validation NLL, serialized bytes, preparation cost, and measured latency, then compare Pareto fronts instead of collapsing objectives into one penalty. Such a design would give diversity-preserving selectors a clearer role than the single-budget objective used here.

# E Appendix: Open Research Questions

- Does tournament's advantage repeat across model families, or is it tied to the Qwen matrix-sensitivity structure?
- At what candidate allowance does tournament first separate reliably from random and Alternis when model and data are held fixed?
- Does Alternis become competitive on a multiobjective Pareto frontier where population diversity has direct value?
- How do learned rotations, second-order reconstruction, or activation-aware scaling alter the mixed-bit fitness landscape?
- Can a cheap proxy preserve tournament's ranking while reducing the cost of direct candidate NLL evaluation?
- Do validation-selected NLL gains transfer to instruction following, coding, retrieval, safety, multilingual data, and long context?
- How should search cost be amortized when a quantized model is personalized to one device or repeatedly deployed to many devices?